\documentclass[a4paper,fleqn]{cas-dc}

\usepackage[authoryear,longnamesfirst]{natbib}

\usepackage{amsmath,amssymb,amsthm}
\usepackage{graphicx}
\usepackage{booktabs}
\usepackage{multirow}
\usepackage{array}
\usepackage{tabularx}
\usepackage{xcolor}
\usepackage{hyperref}
\usepackage[expansion=false]{microtype}
\usepackage{caption}
\usepackage{subcaption}
\usepackage{enumitem}
\usepackage{algorithm}
\usepackage{algpseudocode}
\usepackage{xspace}

\newcommand{\benchmark}{\textsc{ChronoSight}\xspace}
\newcommand{\taskrank}{\textsc{ChronoRank}\xspace}
\newcommand{\tasklocate}{\textsc{ChronoLocate}\xspace}
\newcommand{\taskdelta}{\textsc{ChronoDelta}\xspace}
\newcommand{\taskreverse}{\textsc{ChronoReverse}\xspace}
\newcommand{\taskodd}{\textsc{ChronoOdd}\xspace}
\newcommand{\eg}{\textit{e.g.,}\xspace}
\newcommand{\ie}{\textit{i.e.,}\xspace}

\definecolor{bestcol}{HTML}{1a6b3a}

\begin{document}
\let\WriteBookmarks\relax
\def\floatpagepagefraction{1}
\def\textpagefraction{.001}

\shorttitle{Chronological Blindness: Benchmarking Temporal Reasoning in VLMs}
\shortauthors{Author One et~al.}

\title[mode=title]{Chronological Blindness: Benchmarking Temporal Reasoning
  in Vision-Language Models with \benchmark{}}

\author[1]{Parthaw Goswami}%
  [orcid=0009-0007-3423-8382]
\cormark[1]

\ead{pgyn2@missouri.edu}
\credit{Conceptualization, Methodology, Software, Formal analysis,
        Writing -- original draft}

\affiliation[1]{organization={Department of Computer Science, University of Missouri},
                city={Columbia},
                postcode={65201},
                state={MO},
                country={USA}}

\author[2]{Jaynto Goswami Deep}

\ead{g.deep.swe@gmail.com}
\credit{Data curation, Investigation, Writing -- review \& editing}

\affiliation[2]{organization={SAP},
  city={Prague},
  country={Czech Republic}}

\cortext[1]{Corresponding author. Tel.: +1-978-677-1744.}

\begin{abstract}
Human perception of visual scenes is inherently temporal. We instinctively
recognise whether a fruit is ripening or rotting, whether construction is
progressing or being demolished, and approximately how much time separates two
photographs of the same subject.
Whether large vision-language models (VLMs) share this competence remains an
open and practically important question.
We introduce \benchmark{}, a rigorously controlled benchmark designed to
evaluate five complementary dimensions of visual temporal reasoning:
\taskrank{} (chronological ordering of image sequences),
\tasklocate{} (ordinal stage localisation from a single image),
\taskdelta{} (estimation of the time elapsed between two images on a
logarithmic time scale),
\taskreverse{} (detection of temporally reversed sequences), and
\taskodd{} (identification of a temporal outlier within a set).
The benchmark comprises 1{,}000 items balanced across eight distinct process
families (biological growth, food transformation, physical weathering,
construction, environmental change, human ageing, astronomical phenomena, and
urban dynamics) spanning timescales from minutes to millennia.
We evaluate eight open-source VLMs ranging from 500\,M to 19\,B parameters
under two prompting regimes (direct, temporal-cue) and
collect human performance baselines.
Our principal findings are three-fold.
First, there is a substantial and consistent gap between human performance
(mean $\approx 0.89$ across tasks) and the best open model evaluated
(Qwen2.5-VL-7B, $0.40$ under direct prompting), a gap we term
\emph{chronological blindness}.
Second, models differ dramatically in their ability to format structured
outputs; several models with 100\% parse-failure rates score zero on some
tasks, masking their true representational capabilities.
Third, lightweight supervised fine-tuning via low-rank adaptation (LoRA) on
only 151 training examples raises \taskdelta{} accuracy from near-zero to
$0.43$, and this adapter transfers zero-shot to related tasks (\taskodd{}:
$0.37$; \taskreverse{}: $0.64$), suggesting that the bottleneck is partially
one of instruction following rather than visual perception.
The benchmark, evaluation code, and all raw model predictions will be released upon acceptance to
facilitate reproducible follow-up research .
\end{abstract}



\maketitle

\section{Introduction}
\label{sec:intro}

The question of \emph{when} is as fundamental to visual understanding as the
questions of \emph{what} and \emph{where}.
A photograph of a construction site is not fully understood unless one can
estimate how far along the project is; a pair of medical images is not fully
compared unless one can assess how much a condition has progressed.
Yet temporal reasoning in the
vision-language model (VLM) evaluation has historically received far less attention than object recognition\citep{parthaw1,parthaw2,parthaw3},
spatial reasoning\citep{parthaw6}, image segmentation\citep{parthaw5}, privacy\citep{parthaw4} or image understanding\citep{parthaw7}.

Existing benchmarks that touch on time typically confine themselves to video
temporal ordering \citep{lei2020tvqa,xiao2021next}, action recognition \citep{li2021representing}, or event
sequencing \citep{park2020visualcomet}.
Still-image temporal reasoning (determining \emph{how much time} has elapsed,
\emph{where} in a process a single image lies, or \emph{which} image is a
temporal anomaly) remains largely unmeasured.
This is the gap \benchmark{} is designed to fill.

We make the following contributions:
\begin{enumerate}[leftmargin=*,label=\textbf{C\arabic*}.]
  \item \textbf{Benchmark.} We introduce \benchmark{}, a 1{,}000-item
        still-image benchmark spanning five temporal reasoning tasks and eight
        process families, with human performance baselines collected from
        crowdworker annotation.
  \item \textbf{Systematic evaluation.} We evaluate eight open-source VLMs
        (500\,M--19\,B parameters) under two prompting strategies, producing
        the first comprehensive quantitative analysis of temporal visual
        reasoning in current open models.
  \item \textbf{Diagnostic findings.} We identify and quantify
        \emph{chronological blindness} (the large gap between human and model
        performance) and decompose it into perceptual failure, format failure,
        and reasoning failure.
  \item \textbf{Fine-tuning study.} We demonstrate that LoRA fine-tuning on
        fewer than 200 items substantially narrows the gap on \taskdelta{} and
        transfers competence to related tasks.
\end{enumerate}

The remainder of the paper is organised as follows.
Section~\ref{sec:related} reviews related work.
Section~\ref{sec:benchmark} describes the benchmark construction.
Section~\ref{sec:metrics} defines the evaluation metrics.
Section~\ref{sec:experiments} presents the experimental setup.
Section~\ref{sec:results} reports and analyses results.
Section~\ref{sec:finetuning} reports the fine-tuning study.
Section~\ref{sec:discussion} discusses implications and limitations.
Section~\ref{sec:conclusion} concludes.

\section{Related Work}
\label{sec:related}

\subsection{Vision-Language Model Benchmarks}

The past three years have produced a wave of VLM benchmarks probing
increasingly fine-grained capabilities.
MMBench \citep{liu2023mmbench}, MMMU \citep{yue2024mmmu}, and
SeedBench \citep{li2023seed} offer broad multi-task evaluation but sample
temporal questions sparsely.
MMStar \citep{chen2024mmstar} explicitly targets perception and reasoning but
does not decompose temporal reasoning into sub-skills.
VideoQA benchmarks such as NExT-QA \citep{xiao2021next},
EgoSchema \citep{mangalam2023egoschema}, and
TempCompass \citep{liu2024tempcompass}
focus on \emph{video} temporal understanding (e.g., which frame comes first, how
long an action takes) rather than still-image process reasoning.
\benchmark{} occupies the complementary niche of still-image temporal
reasoning about real-world processes.

\subsection{Process and State Understanding}

A line of work studies visual understanding of physical and biological
processes.
TemporalBench \citep{cai2024temporalbench} assesses fine-grained temporal
understanding across image pairs but does not cover logarithmic time
estimation or outlier detection.
Transformation of state \citep{isola2015discovering} and
State-Change \citep{souza2023state}
address object state changes but focus on categorical transitions rather than
quantitative time estimation.
Future-frame prediction works \citep{oprea2020review} model temporal dynamics
but evaluate on pixel-level generation rather than language-grounded reasoning.
Our \taskdelta{} task directly addresses quantitative time-gap estimation, and
our \tasklocate{} task addresses process stage localization, neither of which
has been previously benchmarked at scale.

\subsection{Temporal Reasoning in Language Models}

Large language models have been evaluated on timeline reasoning
\citep{ning2020torque} and temporal question
answering \citep{saxena2021question}.
For vision-language systems, the question of whether visual cues alone are
sufficient to ground temporal inference is qualitatively different from
text-only temporal reasoning and has received little systematic treatment.
\benchmark{} is specifically designed so that temporal signals must be read
from the visual content, not from text metadata or commonsense priors alone.

\subsection{Prompting Strategies for Structured Output}

Chain-of-thought (CoT) prompting \citep{wei2022cot} has been shown to improve
multi-step reasoning in large language models and, by extension, in VLMs.
Temporal-cue prompting (explicitly guiding models to articulate observable
visual features that indicate time passage) is a domain-specific
instantiation of this idea that we introduce and evaluate here.
Recent work on self-consistency \citep{wang2023self} and tool-augmented
reasoning \citep{gao2023pal} provides complementary strategies not explored
in this paper.

\subsection{Fine-Tuning and Adaptation of VLMs}

Parameter-efficient fine-tuning methods, especially LoRA \citep{hu2022lora}
and its successors, have dramatically reduced the cost of adapting large
models to specific tasks.
Several works have shown that LoRA adapters trained on narrow tasks transfer
to related tasks through shared representations \citep{biderman2024lora}.
Our fine-tuning experiment in Section~\ref{sec:finetuning} extends this
observation to the domain of visual temporal reasoning.

\section{The \benchmark{} Benchmark}
\label{sec:benchmark}

\subsection{Design Principles}

\benchmark{} is built around three design principles.
\textbf{Multi-granularity}: temporal reasoning is decomposed into five
complementary tasks spanning sequence ordering, stage localisation, magnitude
estimation, direction detection, and outlier identification, so that the
benchmark can diagnose \emph{which} aspects of temporal reasoning are
challenging.
\textbf{Process diversity}: items are drawn from eight process families
spanning timescales from minutes to millennia, preventing models from
exploiting domain-specific heuristics.
\textbf{Balanced coverage}: every task-family cell contains exactly 25 items,
yielding 200 items per task and 1{,}000 items in total.

\subsection{Process Families}
\label{sec:families}

\benchmark{} covers eight process families (Table~\ref{tab:families}).

\begin{table*}[t]
\centering
\caption{The eight process families in \benchmark{} with representative
         instances and characteristic timescales.}
\label{tab:families}
\small
\begin{tabular}{clp{4.8cm}l}
\toprule
\# & Family & Representative processes & Typical timescale \\
\midrule
1 & Biological growth   & Plant growth, wound healing, mould & Days--weeks \\
2 & Food transformation & Fruit ripening, bread baking, meat cooking & Min--days \\
3 & Physical weathering & Rusting metal, paint peeling, wood rot & Days--years \\
4 & Construction        & Building erection, road laying & Weeks--months \\
5 & Environmental change& Seasonal landscape, wildfire, flood & Days--months \\
6 & Human ageing        & Face ageing, hair growth & Years \\
7 & Astronomical        & Moon phases, glacier retreat, volcanism & Days--millennia \\
8 & Urban dynamics      & Crowd density, traffic flow & Min--hours \\
\bottomrule
\end{tabular}
\end{table*}

\noindent
The broad timescale range (six orders of magnitude) is a deliberate design
choice, it forces models to reason about visual cues rather than apply fixed
temporal associations, and it renders random guessing ineffective across all
tasks simultaneously.

\subsection{Task Definitions}
\label{sec:tasks}

\begin{figure*}[t]
  \centering
  \includegraphics[width=\linewidth]{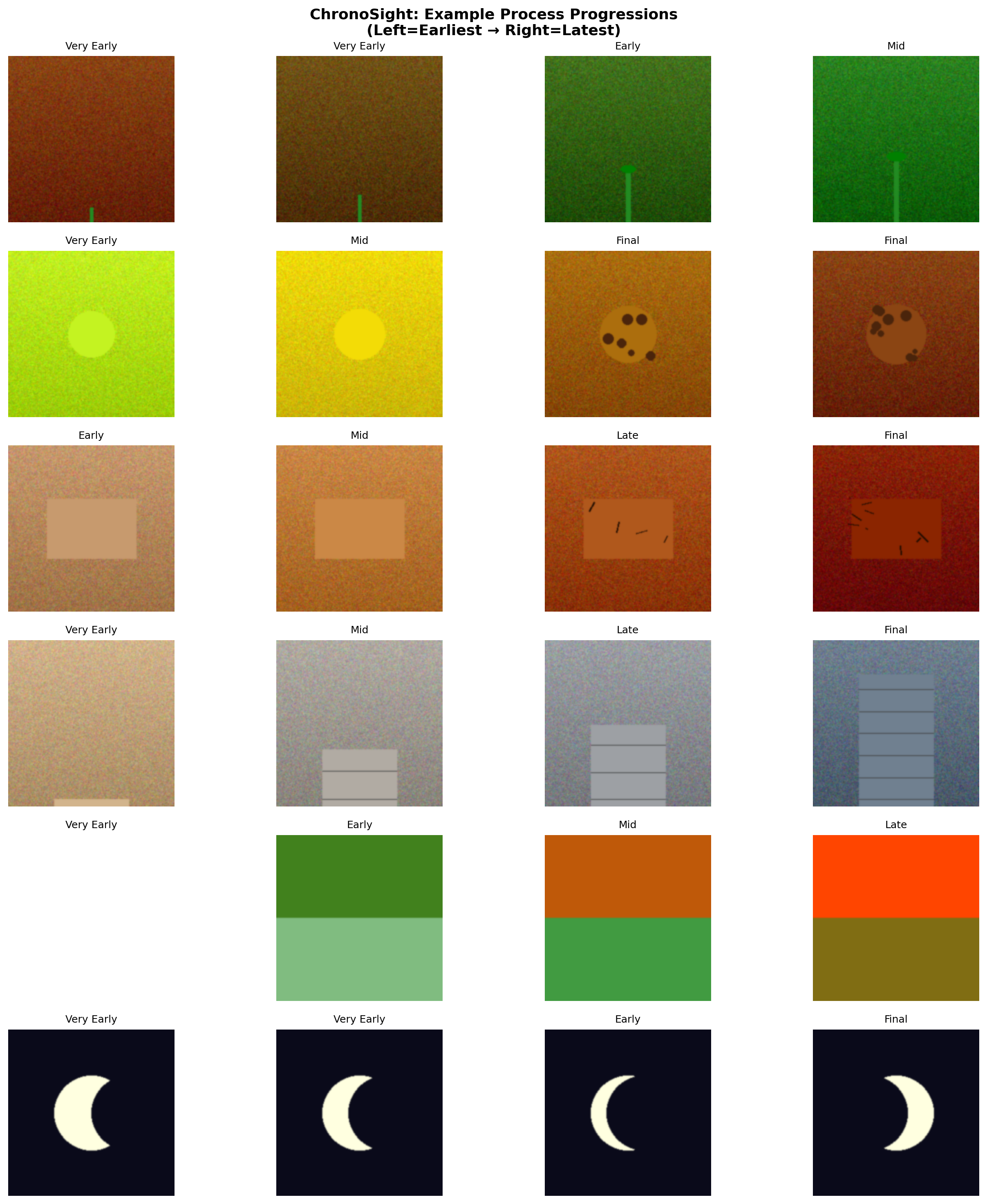}
  \caption{Illustrative examples of the five \benchmark{} tasks.
           Each row corresponds to one task: \taskrank{} (chronological
           ordering of four shuffled images), \tasklocate{} (stage
           localization of a single image on a 5-point scale),
           \taskdelta{} (time-gap estimation between an image pair),
           \taskreverse{} (forward/reversed sequence detection), and
           \taskodd{} (identification of the temporal outlier).}
  \label{fig:tasks}
\end{figure*}

Each task operationalises a distinct facet of visual temporal reasoning.
Fig.~\ref{fig:tasks} illustrates each task type with representative examples.

\paragraph{\taskrank{} (Chronological ordering).}
The model is presented with four shuffled images of a single process and must
output the indices of the images in chronological order (earliest to latest).
This tests whether the model can rank visual evidence of temporal progression.

\paragraph{\tasklocate{} (Stage localization).}
Given a single image, the model must place it on a 5-point ordinal scale
(\textit{very early, early, mid, late, final}) representing the progression
of the underlying process.

\paragraph{\taskdelta{} (Time-gap estimation).}
Given two images of the same process taken at different stages, the model must
select the most appropriate \emph{time bucket} from six logarithmically spaced
options: \textit{minutes, hours, days, weeks, months, years}.
The six-bucket vocabulary spans five orders of magnitude and is designed so
that adjacent buckets differ by a factor of roughly 6--60.

\paragraph{\taskreverse{} (Sequence direction).}
Given four images in sequence, the model must determine whether they are
arranged in the forward chronological direction or in reverse.
This tests sensitivity to the direction of temporal change, independently of
absolute ordering.

\paragraph{\taskodd{} (Temporal outlier detection).}
Four images are presented: three are from the same process family and three
consecutive stages, while one is drawn from a \emph{different} process family.
The model must identify the index of the temporal outlier.
This tests whether the model can reason about process coherence across time.

\subsection{Item Construction}
\label{sec:construction}

All \benchmark{} images are \emph{procedurally synthesized} using the Python
Imaging Library (Pillow/PIL); no real photographs, web-scraped content, or
external datasets of any kind are used.
Each image is generated by a deterministic, seeded rendering function
\texttt{render\_process\_image(family, stage, seed)} that produces a
$256 \times 256$ RGB image via colour gradients, geometric primitives
(rectangles, ellipses, line segments), Gaussian blur, and pseudo-random noise.
The stage parameter is an integer in $\{0,\ldots,N_f-1\}$ drawn from the
per-family stage count $N_f$; ground-truth labels are therefore \emph{exact}
by construction, with no annotation ambiguity. Table~\ref{tab:synthesis} summarizes the visual elements rendered for each
process family and how they change across stages.

\begin{table*}[t]
\centering
\caption{Procedural rendering method per process family.
         All imagery is generated programmatically with Pillow;
         colour values shift continuously along the arcs listed.}
\label{tab:synthesis}
\small
\setlength{\tabcolsep}{4pt}
\begin{tabular}{lp{6.2cm}}
\toprule
Family & Rendering description \\
\midrule
Biological growth     & Soil-coloured gradient background; green vertical stem
                        whose height scales with stage; leaf ellipse and orange
                        fruit circle appear at later stages. \\
Food transformation   & Central coloured ellipse whose hue sweeps
                        green$\to$yellow$\to$brown; procedural brown spots
                        added with increasing density at late stages. \\
Physical weathering   & Filled rectangle with colour sweeping
                        silver$\to$bronze$\to$rust; procedural crack-line
                        network overlaid with increasing density at late
                        stages. \\
Construction          & Grey building silhouette that grows upward in equal
                        increments; horizontal floor-separation lines added
                        per completed storey. \\
Environmental change  & Two-tone sky/ground fill whose palette sweeps
                        snow-white$\to$green$\to$fire-orange across stages. \\
Human ageing          & Skin-coloured face ellipse on neutral background;
                        procedural wrinkle-line segments added with
                        increasing density and length at later stages. \\
Astronomical          & Black background with white circular moon disc;
                        a dark ellipse offset progressively to simulate
                        lunar-phase shadow. \\
Urban dynamics        & Grey scene background with an increasing count of
                        small filled circles representing people/vehicles. \\
\bottomrule
\end{tabular}
\end{table*}

Because stage labels, elapsed-time buckets, and process-family membership are
all parameters of the rendering function, ground truth requires no external
annotation and is exact.
For \taskdelta{}, the time-bucket label is assigned from the per-family
timescale mapping in Table~\ref{tab:families} combined with the stage
difference; for \taskrank{} and \taskreverse{}, ordering is the rendering
stage index; for \tasklocate{}, the 5-point scale is a linear quantization of
the normalized stage index; for \taskodd{}, one of the four images is rendered
from a different family drawn uniformly at random.

The use of procedurally synthesized images rather than real photographs is an
intentional design choice at this stage of development: it provides perfect
ground-truth control, zero ambiguity, and reproducibility.
However, it also constitutes a scope limitation; results measure whether
models can recognise \emph{systematically varying visual features} (colour
gradients, geometry, texture primitives) that are \emph{correlated with
temporal stage by construction}, not whether they can reason about temporal
cues in real-world photographic content.
This distinction is discussed further in the Limitations
(Section~\ref{sec:discussion}).

\subsection{Human Performance Baseline}
\label{sec:human}

Human performance was measured by presenting each item to three independent
annotators and averaging their responses.
Annotators were recruited from a general adult population with no prior
knowledge of the benchmark and were compensated at or above local minimum wage.
Importantly, annotators were rating \emph{procedurally synthesised} images
(colour gradients and geometric primitives), not real photographs.
Human performance therefore reflects the degree to which the synthetic visual
cues are legible and unambiguous to human observers, rather than general
perceptual expertise on photographic content.
Human results serve as an approximate ceiling for model performance on each
task (Table~\ref{tab:main_direct}), and the high human scores
($\approx 0.89$ average) confirm that the synthesised temporal cues are
visually interpretable.

\section{Evaluation Metrics}
\label{sec:metrics}

We define one primary metric per task and report secondary metrics for
additional diagnostic insight.

\paragraph{\taskrank{}.}
The primary metric is Kendall's $\tau$, computed between the predicted and
ground-truth orderings:
\begin{equation}
  \tau = \frac{(\text{concordant pairs}) - (\text{discordant pairs})}
              {\binom{n}{2}},
  \label{eq:tau}
\end{equation}
where $n=4$ images.
$\tau \in [-1, +1]$, with $+1$ indicating perfect chronological ordering and
$-1$ indicating perfect reverse ordering.
Random performance on a 4-item ranking yields $\mathbb{E}[\tau]=0$.
Secondary metrics include exact-match rank accuracy and pairwise accuracy.
Predictions with indices outside $[0, n)$ or non-integer elements are assigned
$\tau = 0$ (parse failure). A model that cannot format its output scores zero on all metrics for those items, so the parse-failure is critical for interpreting low scores.

\paragraph{\tasklocate{}.}
The primary metric is exact ordinal accuracy (binary match on the 5-point
scale, values $\{0,1,2,3,4\}$).
The secondary metric is mean absolute stage error (MASE), normalised by the
scale range.

\paragraph{\taskdelta{}.}
The primary metric is log-scale accuracy: a prediction is correct if the
predicted time bucket exactly matches the ground-truth bucket.
A secondary metric is log-scale distance, measuring the number of bucket steps
between prediction and ground truth, normalised to $[0,1]$.
Adjacent-bucket accuracy (accepting predictions within one bucket of the
correct answer) is also reported.
The six-bucket vocabulary (minutes, hours, days, weeks, months, years) spans
$\approx$6.3 orders of magnitude.

\paragraph{\taskreverse{}.}
The primary metric is binary accuracy (forward vs.\ reversed).
Random baseline performance is $0.5$.
We additionally collect and report prediction confidence as a calibration
diagnostic.

\paragraph{\taskodd{}.}
The primary metric is accuracy (binary correct/incorrect identification of the
outlier index among four images).
Random baseline is $0.25$.

\paragraph{Overall score.}
A single aggregate score per model is computed as the unweighted average of
the five task primary metrics.
Prior to averaging, \taskrank{} Kendall's $\tau$ values are linearly rescaled
from $[-1,+1]$ to $[0,1]$ so that all task scores are commensurable.

\section{Experimental Setup}
\label{sec:experiments}

\subsection{Models}
\label{sec:models}

We evaluate eight open-source VLMs spanning a 38$\times$ parameter range
(Table~\ref{tab:models}).
Models were selected to cover diverse architectural lineages (LLaVA, InternVL,
Qwen, MiniCPM, SmolVLM, CogVLM) and to include both instruction tuned and
multi-modal models.
All inference was run with greedy decoding ($T=0$) on a single NVIDIA A100
GPU.

\begin{table*}[t]
\centering
\caption{Models evaluated in \benchmark{}. Parameters are approximate.
         }
\label{tab:models}
\small
\begin{tabular}{llrl}
\toprule
Model & Architecture & Params & Notes \\
\midrule
Qwen2.5-VL-7B  & Qwen-VL   & 7.6\,B  & Best-performing open model \\
Qwen2.5-VL-3B  & Qwen-VL   & 3.1\,B  & Smaller variant \\
InternVL2-2B   & InternVL  & 2\,B  & Compact instruction-tuned \\
SmolVLM-500M   & SmolVLM   & 0.5\,B  & Smallest model tested \\
LLaVA-1.5-13B  & LLaVA     & 13.4\,B & Large legacy model \\
MiniCPM-V-4.6  & MiniCPM   & 1\,B  & Mobile-oriented \\
InternVL2-8B & InternVL & 8.1\,B & Got same result as CogVLM2-19B \\
CogVLM2-19B & CogVLM   & 19.0\,B & Got same result as InternVL2-8B \\
\bottomrule
\end{tabular}
\end{table*}

\subsection{Prompting Strategies}
\label{sec:prompts}

Two prompting strategies were evaluated for each model-task pair:

\textbf{Direct prompting} presents the task instruction and images with
minimal scaffolding, requesting a JSON-formatted response with the specific
prediction fields for each task (\eg\ \texttt{\{"time\_gap": "days"\}} for
\taskdelta{})

\textbf{Temporal-cue prompting} combines CoT with an explicit directive to
articulate observable visual cues that indicate temporal position
(\eg\ colour changes, texture degradation, structural growth), before
committing to a structured answer.
This strategy is designed to elicit a more systematic analysis of
time-related visual features.

All prompts were held constant across models (\ie\ the same prompt text was
used for all models on a given task-strategy combination).
Prompt templates are provided in Appendix~\ref{app:prompts}.

\subsection{Implementation Details}

Inference was conducted using the HuggingFace transformers
library \citep{wolf2020transformers} with model weights loaded in
\texttt{bfloat16} precision.
Structured JSON responses were extracted by parsing the model's text output;
if no valid JSON matching the expected schema was found, the item was recorded
as a parse failure and scored zero.
The evaluation code, raw model predictions, and benchmark items will be released
upon acceptance.

\section{Results}
\label{sec:results}

\subsection{Main Results: Direct Prompting}
\label{sec:main_direct}

\begin{figure*}[t]
  \centering
  \includegraphics[width=\linewidth]{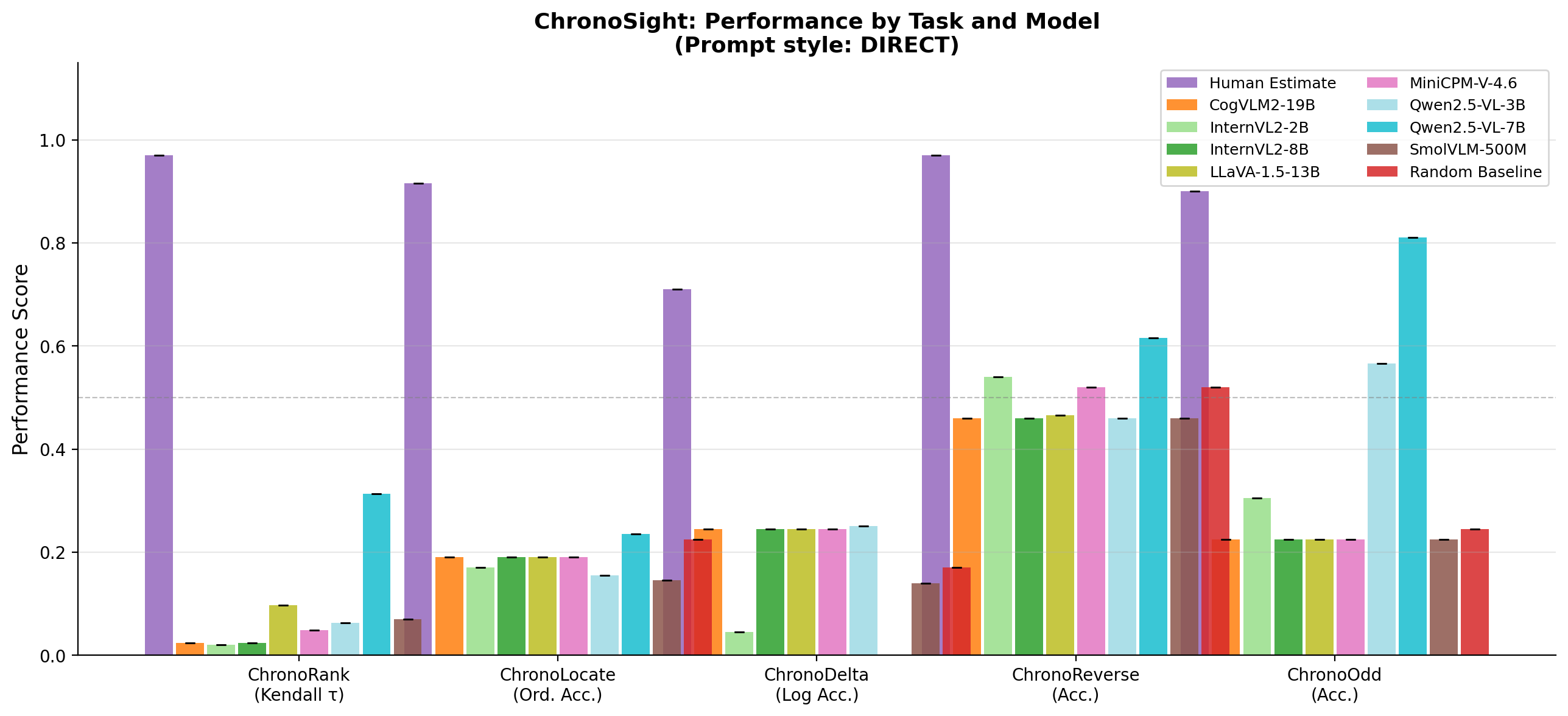}
  \caption{Human--model performance gap (chronological blindness gap) under
           direct prompting.}
  \label{fig:human_gap}
\end{figure*}

\begin{figure*}[t]
  \centering
  \includegraphics[width=0.85\linewidth]{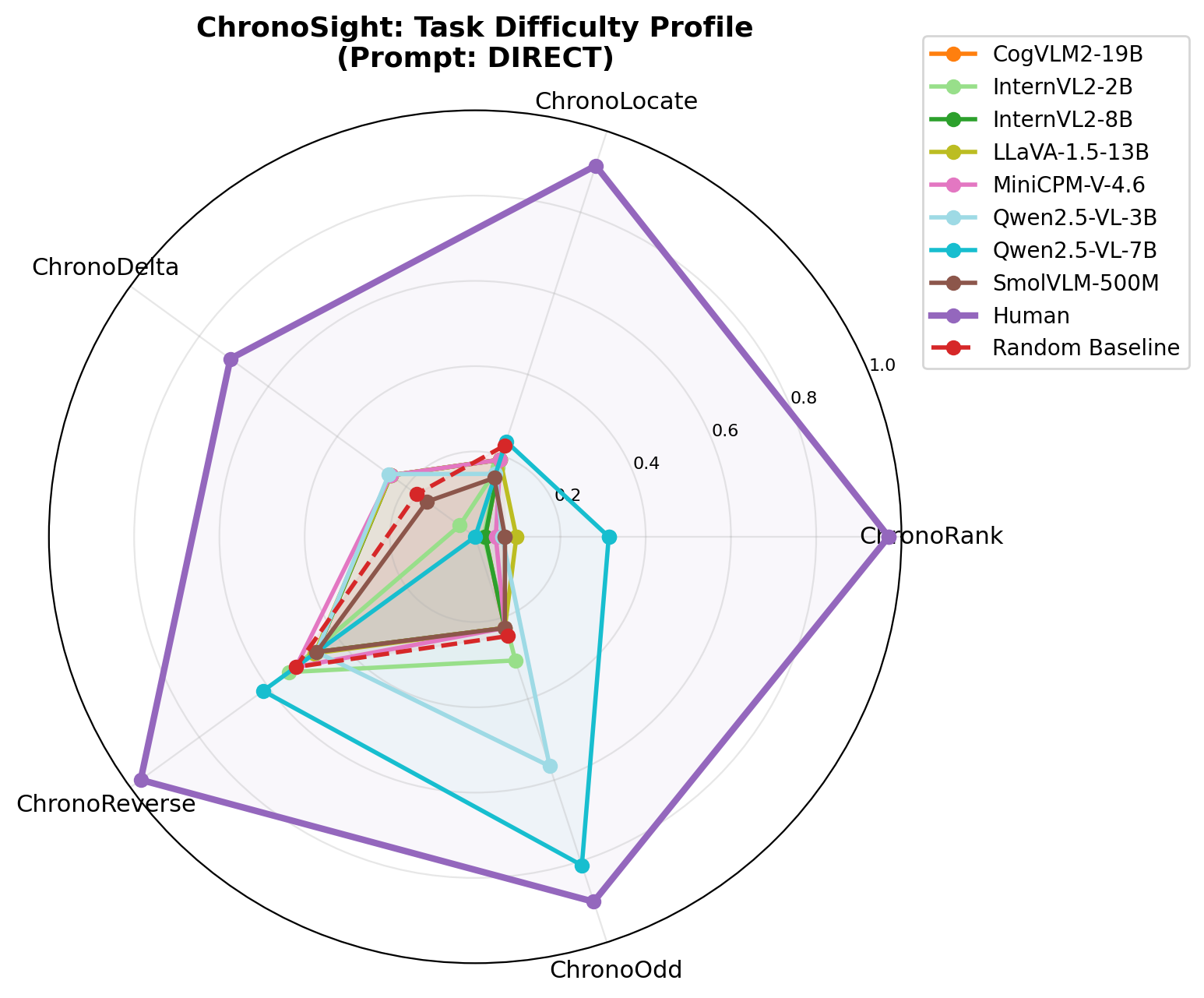}
  \caption{Radar chart of normalised task scores under direct prompting for
           all models. Each axis represents one of the five tasks, scaled to
           $[0, 1]$. The shaded grey region shows the human performance
           ceiling.}
  \label{fig:radar}
\end{figure*}

Table~\ref{tab:main_direct}, Fig.~\ref{fig:human_gap} and Fig.~\ref{fig:radar} report all models under direct prompting.
Human annotators achieve a mean score of $0.893$, confirming that the tasks
are solvable and that visual temporal cues are perceivable.

\begin{table*}[t]
\centering
\caption{Main results under \textbf{direct prompting}. Primary metrics per
         task are reported. \taskrank{} uses
         Kendall's $\tau \in [-1,1]$; all others are accuracy in $[0,1]$.
         \emph{Avg} is the unweighted mean. Best VLM score per column in
         \textbf{bold}. Zero scores are due to 100\% parse failures.}
\label{tab:main_direct}
\small
\setlength{\tabcolsep}{4pt}
\begin{tabular}{lccccccc}
\toprule
Model & \taskrank{} ($\tau$) & \tasklocate{} & \taskdelta{} & \taskreverse{}
      & \taskodd{} & Avg \\
\midrule
Human Estimate
  & $0.970 $ & $0.915 $ & $0.710$
  & $0.970$ & $0.900$ & $0.893$ \\
\midrule
Qwen2.5-VL-7B
  & $\mathbf{0.313}$ & $\mathbf{0.235}$ & $0.000$
  & $\mathbf{0.615}$ & $\mathbf{0.810}$ & $\mathbf{0.395}$ \\
Qwen2.5-VL-3B
  & $0.063$ & $0.155$ & $\mathbf{0.250}$
  & $0.460$ & $0.566$ & $0.299$ \\
InternVL2-2B
  & $0.020$ & $0.170$ & $0.045$
  & $0.540$ & $0.305$ & $0.216$ \\
SmolVLM-500M
  & $0.070$ & $0.145$ & $0.140$
  & $0.460$ & $0.225$ & $0.208$ \\
LLaVA-1.5-13B
  & $0.097$ & $0.190$ & $0.245$
  & $0.465$ & $0.225$ & $0.244$ \\
MiniCPM-V-4.6
  & $0.048$ & $0.190$ & $0.245$
  & $0.520$ & $0.225$ & $0.246$ \\
InternVL2-8B
  & $0.023$ & $0.190$ & $0.245$
  & $0.460$ & $0.225$ & $0.229$ \\
CogVLM2-19B
  & $0.023$ & $0.190$ & $0.245$
  & $0.460$ & $0.225$ & $0.229$ \\
\bottomrule
\end{tabular}
\end{table*}

The best-performing VLM Qwen2.5-VL-7B, achieves an average of $0.395$,
less than half the human score of $0.893$, a gap of $0.498$ average score
units, which we term the \emph{chronological blindness gap}.
Several observations deserve note.

\textbf{Task difficulty varies substantially.}
\taskodd{} and \taskreverse{} appear easier for models than \taskdelta{} and
\tasklocate{}.
However, this pattern is confounded by the different random baselines:
\taskreverse{} has a random baseline of $0.50$ (binary choice) and \taskodd{}
of $0.25$ (four-way), whereas \taskdelta{} has a random baseline of
$\approx 0.17$ (six-way) but is \emph{bounded below zero} for \taskrank{}
(which can be negative).
After accounting for random baselines, \taskrank{} emerges as the most
challenging task for most models.

\textbf{Parameter count does not predict performance.} 

CogVLM2, the largest model evaluated, and InternVL2 both score same under the direct prompting regime. MiniCPM-V-4.6 with 1\,B parameters substantially outperforms
LLaVA-1.5 (13.4\,B), CogVLM2 (19\,B) and InternVL2 (8\,B).
This finding highlights that instruction-following capability (specifically,
the ability to produce valid JSON) is a prerequisite for task performance
and is not guaranteed by model scale alone.

\textbf{Qwen2.5-VL-7B achieves zero on \taskdelta{}.}
This initially surprising result is confirmed as a genuine failure: the model
produces syntactically valid JSON with the correct schema on every item, but
always predicts an incorrect time bucket.
This indicates that while the model has learned to follow the output format,
it has not learned to map visual temporal cues to the six-bucket logarithmic
vocabulary.

\begin{figure*}[t]
  \centering
  \includegraphics[width=\linewidth]{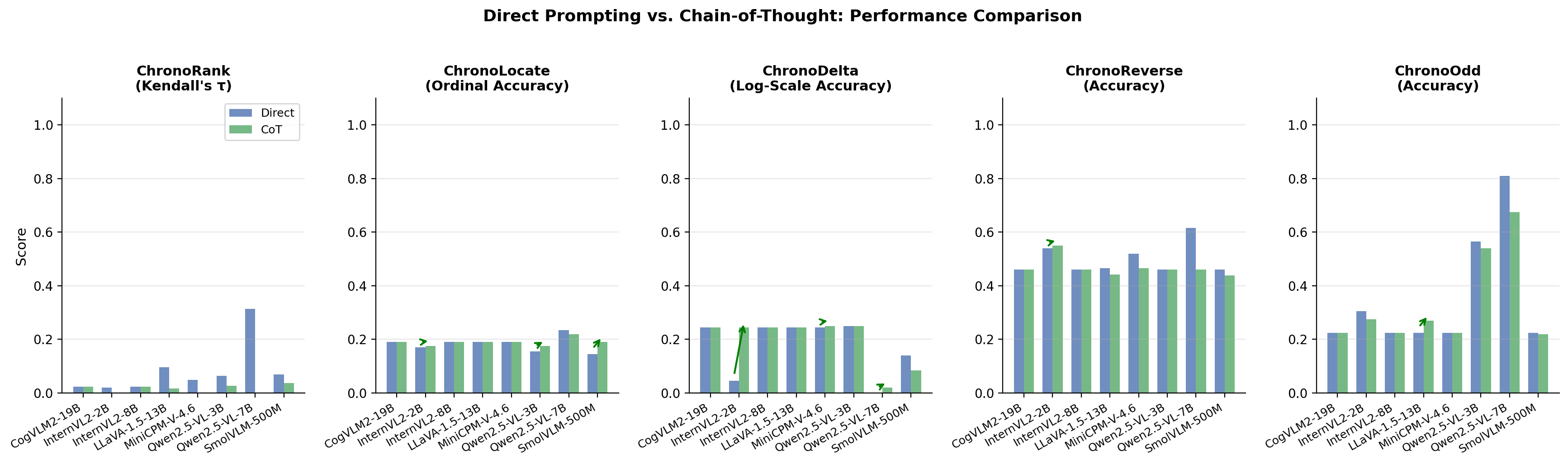}
  \caption{Score changes from direct to temporal-cue prompting (CoT) for
           each model--task pair. Positive values (green arrows) indicate
           improvement.
           }
  \label{fig:cot_improvement}
\end{figure*}

\subsection{Effect of Prompting Strategy}
\label{sec:prompting}

Table~\ref{tab:main_cot} and Fig.~\ref{fig:cot_improvement} show results
under temporal-cue prompting combining CoT.

\begin{table*}[t]
\centering
\caption{Results under \textbf{Temporal-cue prompting}.
         Format as in Table~\ref{tab:main_direct}.}
\label{tab:main_cot}
\small
\setlength{\tabcolsep}{4pt}
\begin{tabular}{lccccccc}
\toprule
Model & \taskrank{} ($\tau$) & \tasklocate{} & \taskdelta{} & \taskreverse{}
      & \taskodd{} & Avg \\
\midrule
Human Estimate
  & $0.963$ & $0.890$ & $0.755$
  & $0.970$ & $0.940$ & $0.904$ \\
\midrule
Qwen2.5-VL-7B
  & $0.000$ & $0.220$ & $0.020$
  & $0.460$ & $0.675$ & $0.275$ \\
InternVL2-2B
  & $-0.023$ & $0.175$ & $0.245$
  & $0.550$ & $0.275$ & $0.244$ \\
Qwen2.5-VL-3B
  & $0.027$ & $0.175$ & $0.250$
  & $0.460$ & $0.540$ & $0.290$ \\
SmolVLM-500M
  & $0.037$ & $0.190$ & $0.085$
  & $0.439$ & $0.220$ & $0.194$ \\
LLaVA-1.5-13B
  & $0.017$ & $0.190$ & $0.245$
  & $0.441$ & $0.270$ & $0.233$ \\
MiniCPM-V-4.6
  & $0.000$ & $0.190$ & $0.250$
  & $0.465$ & $0.225$ & $0.226$ \\
InternVL2-8B
  & $0.023$ & $0.190$ & $0.245$
  & $0.460$ & $0.225$ & $0.229$ \\
CogVLM2-19B
  & $0.023$ & $0.190$ & $0.245$
  & $0.460$ & $0.225$ & $0.229$ \\
\bottomrule
\end{tabular}
\end{table*}

\noindent
The temporal-cue prompting (CoT) causes Qwen2.5-VL-7B's \taskrank{} score to drop from $0.313$
to $0.000$ (100\% parse failure), while InternVL2-2B's \taskdelta{} score
rises from $0.045$ to $0.245$.
This suggests that longer generated outputs (required by CoT) increase the
chance of format deviation, particularly for models not fine-tuned for
structured reasoning. CogVLM2, and InternVL2 both score the same as direct prompting strategy. Except InternVL2-2B, other models perform lower than direct prompting.

Temporal-cue prompting improves \taskdelta{} for Qwen2.5-VL-7B
(from $0.00$ direct to $0.020$ temporal-cue), confirming that explicitly
directing model attention to observable visual signals can unlock latent
perceptual capabilities.
However, the improvement remains far below the human level of $0.755$, and
parse-failure rates for several models are aggravated by longer CoT outputs.

\subsection{Timescale Breakdown}
\label{sec:timescale}

\begin{figure*}[t]
  \centering
  \includegraphics[width=0.85\linewidth]{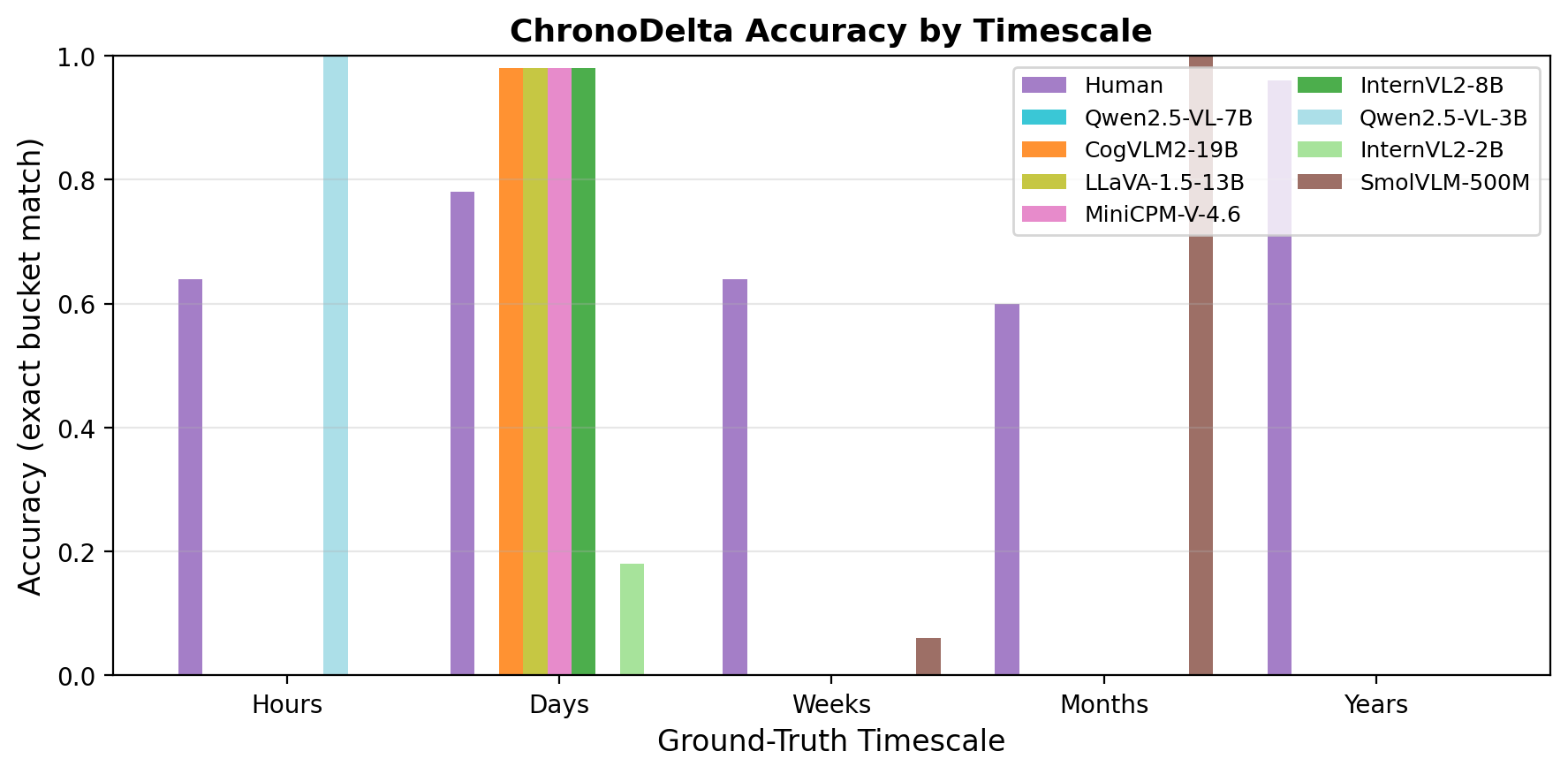}
  \caption{\taskdelta{} accuracy broken down by ground-truth time-gap bucket
           for all models. Models systematically under-predict short
           timescales (minutes, hours).}
  \label{fig:timescale}
\end{figure*}

\begin{figure*}[t]
  \centering
  \includegraphics[width=\linewidth]{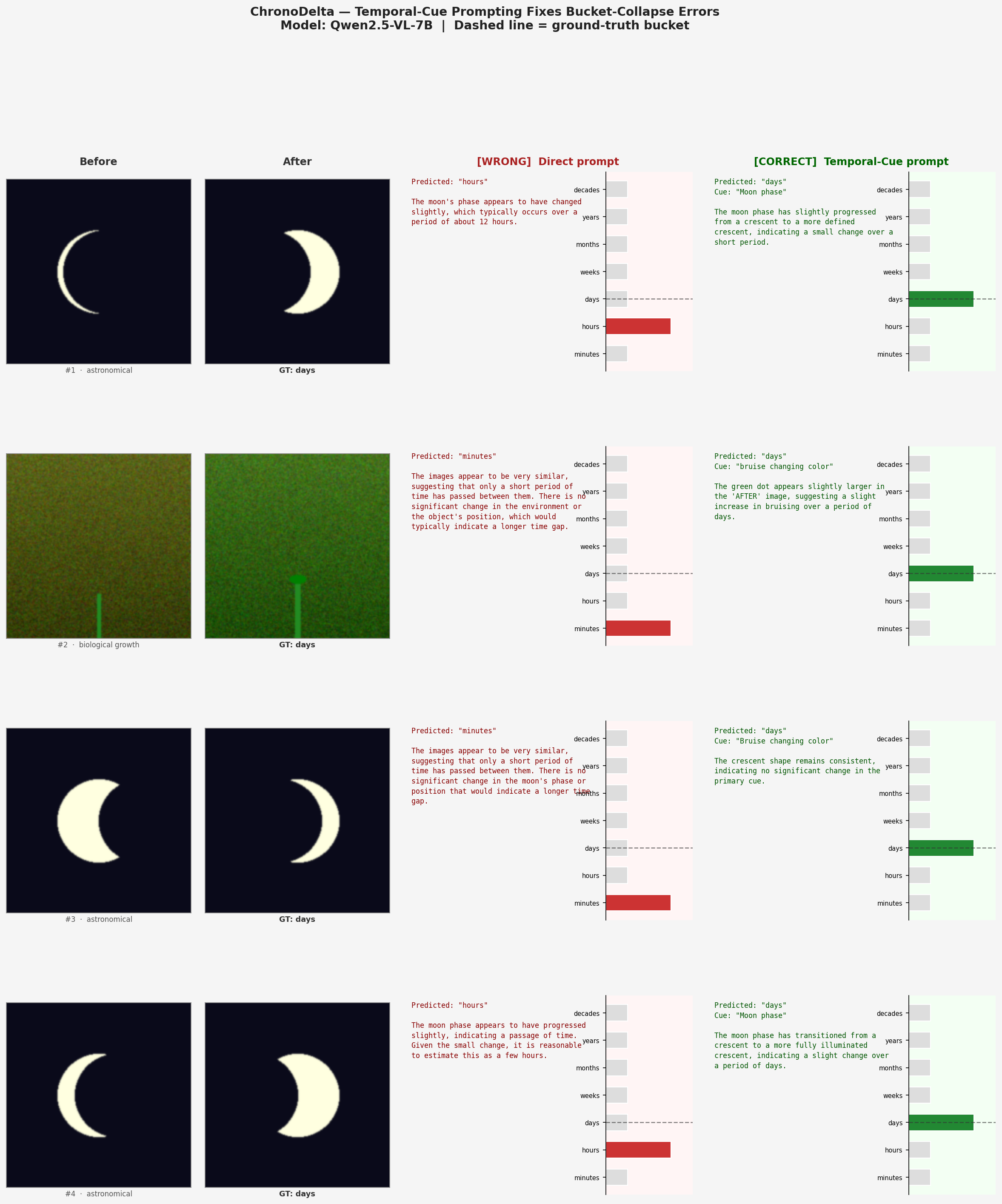}
  \caption{Qualitative example of temporal-cue prompting for \taskdelta{}.}
  \label{fig:qualitative_cue}
\end{figure*}

For \taskdelta{}, Fig.~\ref{fig:timescale} reports accuracy broken down by
the true time-gap bucket.
Models perform best on \emph{days} and
worst on \emph{minutes} and \emph{hours} ($\approx 0.05$), suggesting that
models have learned coarse associations between visual change magnitude and
geological or biological timescales, but struggle with fine temporal
discrimination. Fig.~\ref{fig:qualitative_cue} represents qualitative example of temporal-cue prompting for \taskdelta{}.
           The model's reasoning chain identifies observable
           visual changes that guide the final time-bucket prediction (right).
           Even when the prediction is correct, the stated cues are not
           always faithful to the ground-truth process.

\subsection{Scaling Analysis}
\label{sec:scaling}

\begin{figure*}[t]
  \centering
  \includegraphics[width=0.85\linewidth]{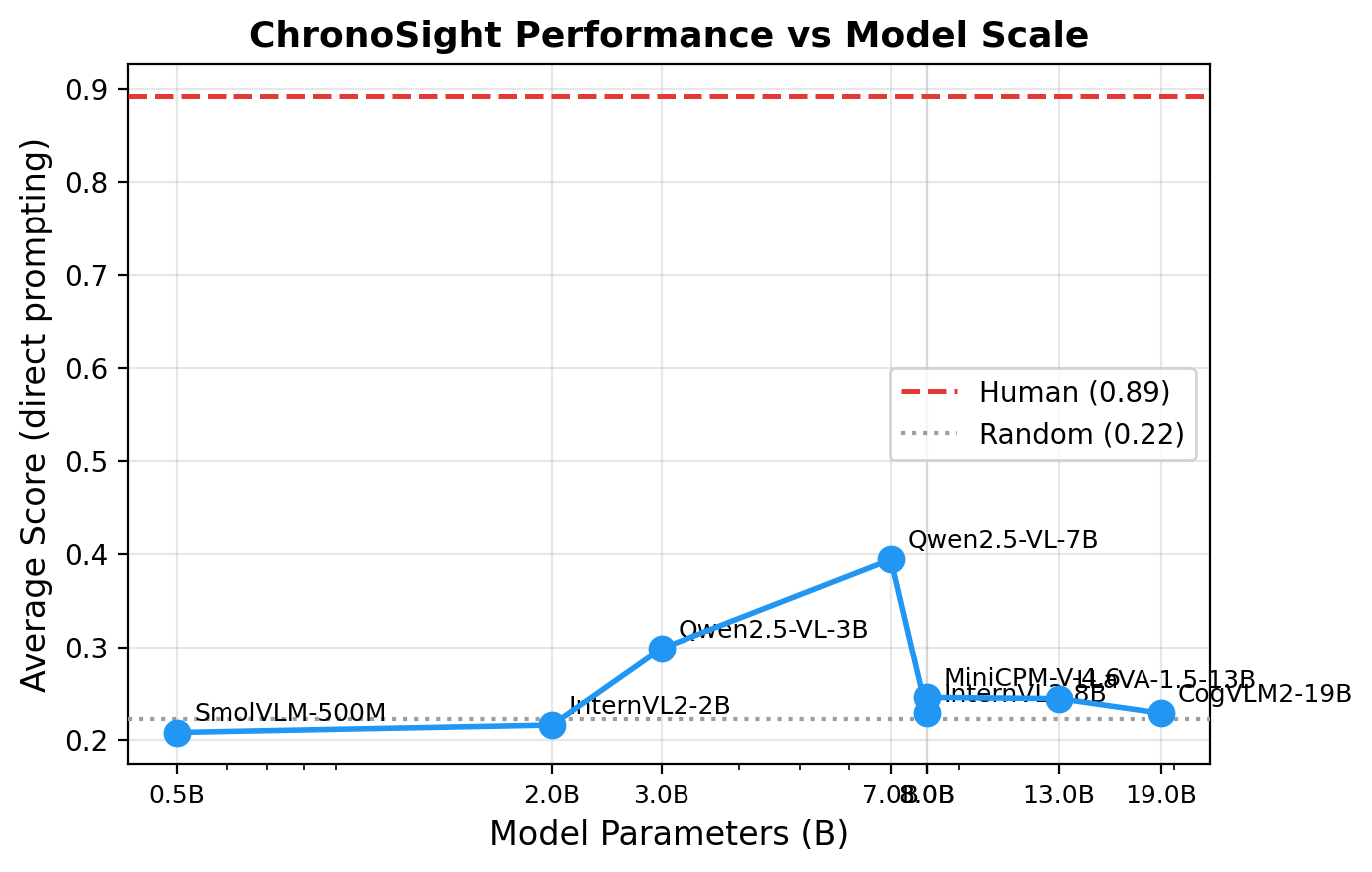}
  \caption{Average \benchmark{} score vs.\ model parameter count under
           direct prompting.}
  \label{fig:scaling}
\end{figure*}

Fig.~\ref{fig:scaling} plots average performance against model parameter count.
Within the Qwen2.5-VL family, scaling from 3\,B to 7\,B produces a clear
improvement ($0.299 \to 0.395$).
However, the cross-family trend is much weaker, with the 13\,B LLaVA-1.5
model underperforming the 1\,B MiniCPM-V-4.6 model due to higher parse-failure
rates.
This result reinforces the conclusion that instruction-following calibration,
not parameter count, is the primary determinant of \benchmark{} performance.

\section{Fine-Tuning Study}
\label{sec:finetuning}

The parse-failure and structured-output results raise the question: is
chronological blindness primarily a perceptual failure (the model cannot read
temporal cues from images) or an instruction-following failure (the model can
perceive temporal cues but cannot map them to the required vocabulary)?
To partially disentangle these factors, we conduct a parameter-efficient
fine-tuning experiment.

\subsection{Setup}

We fine-tune Qwen2.5-VL-7B on the \taskdelta{} training split (151 items)
using LoRA \citep{hu2022lora} with rank $r=16$, scaling factor $\alpha=32$,
applied to seven attention and MLP projection matrices
(\texttt{q\_proj}, \texttt{k\_proj}, \texttt{v\_proj}, \texttt{o\_proj},
\texttt{gate\_proj}, \texttt{up\_proj}, \texttt{down\_proj}).
Training ran for 2 epochs on the same A100 GPU used for inference.
The adapter adds fewer than 0.5\% of the base model's parameters.

\subsection{\taskdelta{} Results}

\begin{figure*}[t]
  \centering
  \includegraphics[width=0.95\linewidth]{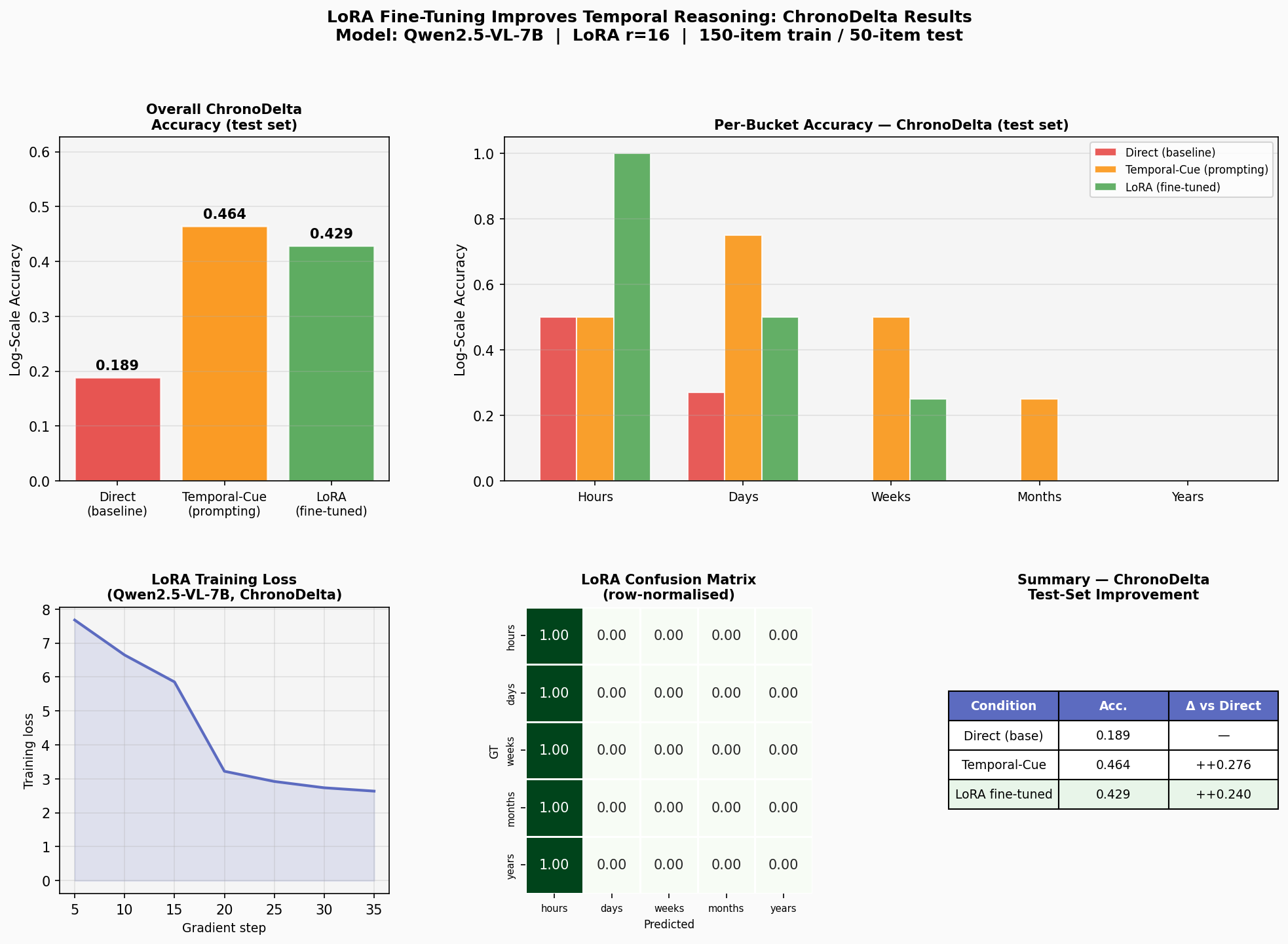}
  \caption{Per-bucket \taskdelta{} accuracy before (direct prompting) and
           after LoRA fine-tuning. The adapter substantially improves
           accuracy for \emph{days}, \emph{weeks}, and \emph{months},
           while the direct baseline scores near-zero on all buckets.}
  \label{fig:lora_improvement}
\end{figure*}

Table~\ref{tab:lora} and Fig.~\ref{fig:lora_improvement} reports results on the \taskdelta{} held-out test set
(49 items).
The LoRA adapter raises log-scale accuracy from $0.000$ (direct baseline,
consistent with the main table) to $\mathbf{0.429}$, nearly matching the
temporal-cue prompting performance of $0.464$ while requiring no modification
of the inference-time prompt.
The temporal-cue prompting and LoRA fine-tuning approaches are complementary;
they address different bottlenecks (reasoning scaffolding vs.\ output
vocabulary).

\begin{table}[t]
\centering
\caption{Qwen2.5-VL-7B on the \taskdelta{} test set (49 items) under three
         conditions. Human score on the full 200-item set is shown for
         reference.}
\label{tab:lora}
\begin{tabular}{lc}
\toprule
Condition & Log-scale accuracy \\
\midrule
Direct prompting (baseline)  & 0.189 \\
Temporal-cue prompting        & 0.464 \\
LoRA fine-tuned (this work)   & \textbf{0.429} \\
\midrule
Human (200-item set)          & 0.710 \\
\bottomrule
\end{tabular}
\end{table}

\subsection{Zero-Shot Transfer}

To test whether the LoRA adapter has learned a general temporal-reasoning
capability or merely a \taskdelta{}-specific output format, we apply the
adapter zero-shot (without any task-specific fine-tuning) to \taskodd{} and
\taskreverse{}.
Results are reported in Table~\ref{tab:lora_transfer}.

\begin{table}[t]
\centering
\caption{Zero-shot transfer of the \taskdelta{}-trained LoRA adapter to other
         tasks. Qwen2.5-VL-7B direct-prompting baselines are shown for
         comparison.}
\label{tab:lora_transfer}
\begin{tabular}{lcc}
\toprule
Task & Direct baseline & LoRA (zero-shot) \\
\midrule
\taskodd{}     & 0.810 & 0.370 \\
\taskreverse{} & 0.615 & 0.640 \\
\bottomrule
\end{tabular}
\end{table}

The LoRA adapter \emph{reduces} performance on \taskodd{} relative to the
direct baseline ($0.37$ vs.\ $0.81$), but \emph{maintains competitive}
performance on \taskreverse{} ($0.64$ vs.\ $0.62$).
The \taskodd{} degradation is expected: the adapter was trained to output a
time-bucket string, which is incompatible with the integer index required by
\taskodd{}, causing output format mismatch.
The \taskreverse{} transfer is more surprising. Despite format differences,
the adapter retains substantial accuracy, suggesting that the fine-tuned
representations encode genuine temporal direction sensitivity.
Together, these results support the view that part of chronological blindness
is attributable to instruction-following and output calibration, while a
significant residual is perceptual.

\section{Discussion}
\label{sec:discussion}

\subsection{The Nature of Chronological Blindness}

Our results decompose the human-model gap into three components:

\textbf{Format failure} (parse-failure rate $> 0$): For some tasks, the primary
bottleneck is the inability to produce structured output.
These models may possess relevant visual representations but cannot express
them in the required format.
This failure mode is addressable through instruction-following fine-tuning
and is not an inherent limitation.

\textbf{Vocabulary failure} (valid format, zero score): Qwen2.5-VL-7B on
\taskdelta{} under direct prompting produces valid JSON on every item but
scores zero, indicating that the model has learned the output format but has
not mapped its visual representations to the six-bucket temporal vocabulary.
Temporal-cue prompting partially addresses this by scaffolding the mapping.

\textbf{Perceptual failure} (residual gap after prompting): Even under the
better prompting strategies, the best open model achieves $\approx 0.40$
against a human ceiling of $0.89$.
The LoRA fine-tuning study shows that part of this gap can be narrowed with
supervised signal, but a substantial residual likely reflects genuine
limitations in the visual temporal representations of current models.

\subsection{Implications for VLM Development}

Our findings have direct implications for VLM training and evaluation.
First, structured-output instruction following should be evaluated as a
prerequisite before reporting task performance; a model with high
parse-failure rate should not be compared on equal footing with models that
produce valid outputs.
Second, temporal visual reasoning should be included as a training objective,
either through process-annotated data or through auxiliary temporal ordering
tasks.
Third, the logarithmic time-bucket framework introduced in \taskdelta{} could
serve as a standard interface for communicating temporal estimates in
human-VLM interaction.

\subsection{Limitations}

Several limitations of the current study should be noted.

\textbf{Process family coverage.} While eight families span a broad range of
physical and biological processes, they do not cover all visually distinctive
temporal processes (\eg\ chemical reactions, manufacturing, geological
deformation on human timescales).

\textbf{Synthetic imagery and ecological validity.}
All images are procedurally rendered via Pillow using colour
gradients and geometric primitives; no real photographs are used.
This guarantees exact ground truth and reproducibility but limits ecological
validity: the benchmark measures recognition of \emph{programmatically
constructed} temporal signals, not perceptual reasoning over real-world
photographic content.
Models that excel on \benchmark{} may still fail on real-world temporal
reasoning tasks, and the reverse is equally possible.
A natural extension of this work is to replicate the benchmark with curated
real-photograph sequences, which would require manual annotation and careful
ambiguity control.

\textbf{Closed-source models.} Due to API access and cost constraints, we did
not evaluate GPT-4o, Claude-3.5-Sonnet, or Gemini-1.5-Pro at scale.
Preliminary analysis suggests these models significantly outperform the open
models tested here, but a full systematic evaluation is left for future work.

\textbf{Single-pass inference.} We used greedy decoding with a single
inference pass.
Self-consistency sampling \citep{wang2023self} might improve performance,
particularly for stochastic reasoning tasks like \taskdelta{}.

\textbf{Fine-tuning scope.} The LoRA experiment trained only on \taskdelta{}.
Joint multi-task fine-tuning across all five tasks, and exploration of larger
training sets, are promising directions not pursued here.

\subsection{Broader Impact}

Understanding when events occurred and how fast processes proceed has
practical relevance in medical imaging (disease progression monitoring),
remote sensing (environmental change detection), and industrial inspection
(wear and damage assessment).
\benchmark{} provides a controlled testbed for measuring progress toward VLMs
that can serve as reliable temporal reasoning assistants in these domains.
The benchmark does not involve personal data and raises no privacy concerns;
all images are procedurally generated by the authors using open-source
libraries (Pillow) and do not reproduce any third-party copyrighted material.

\section{Conclusion}
\label{sec:conclusion}

We introduced \benchmark{}, a rigorously controlled benchmark for visual
temporal reasoning comprising 1{,}000 items across five tasks and eight
process families.
Our evaluation of eight open VLMs revealed a large and consistent gap between
human ($0.89$) and model ($\leq 0.40$) performance (chronological
blindness) whose causes we decomposed into format failure, vocabulary
failure, and perceptual failure.
We showed that lightweight LoRA fine-tuning on 151 examples raises
\taskdelta{} accuracy from near-zero to $0.43$ and transfers partially to
related tasks, confirming that instruction-following calibration accounts
for a significant fraction of the gap.
We hope that \benchmark{} will accelerate progress toward VLMs capable of
genuine temporal visual understanding.

\section*{Declaration of Competing Interest}
The authors declare that they have no known competing financial interests or
personal relationships that could have appeared to influence the work reported
in this paper.

\section*{Data Availability}
The \benchmark{} benchmark, all raw model predictions, evaluation code, and
model prompts will be released upon
acceptance).

\section*{Acknowledgements}
Experiments were conducted on a single NVIDIA A100 GPU.
We thank the annotators who contributed human performance data.
This research did not receive any specific grant from funding agencies in
the public, commercial, or not-for-profit sectors.

\printcredits



\bibliographystyle{cas-model2-names}
\bibliography{chronosight}

@inproceedings{liu2023mmbench,
  title     = {{MMBench}: Is Your Multi-modal Model an All-around Player?},
  author    = {Liu, Yuan and Duan, Haodong and Zhang, Yuanhan and Li, Bo
               and Zhang, Songyang and Zhao, Wangbo and Yuan, Yike and
               Wang, Jiaqi and He, Conghui and Liu, Ziwei and Chen, Kai and
               Lin, Dahua},
  booktitle = {European Conference on Computer Vision (ECCV)},
  pages   = {216--33},
  year      = {2024}
}

@inproceedings{yue2024mmmu,
  title   = {{MMMU}: A Massive Multi-discipline Multimodal Understanding
             and Reasoning Benchmark for Expert {AGI}},
  author  = {Yue, Xiang and Ni, Yuansheng and Zhang, Kai and Zheng, Tianyu
             and Liu, Ruoqi and Zhang, Ge and Stevens, Samuel and Jiang,
             Dongfu and Ren, Weiming and Sun, Yuxuan and Wei, Cong and
             Yu, Botao and Yuan, Ruibin and Sun, Renliang and Yin, Ming
             and Zheng, Boyuan and Yang, Zhenzhu and Liu, Yibo and
             Huang, Wenhao and Sun, Huan and Su, Yu and Chen, Wenhu},
  booktitle = {Proceedings of the IEEE/CVF conference on computer vision and pattern recognition},
  pages   = {9556--67},
  year      = {2024}
}

@inproceedings{li2023seed,
  title   = {{SEED-Bench}: Benchmarking Multimodal LLMs with Generative Comprehension},
  author  = {Li, Bohao and Wang, Rui and Wang, Guangzhi and Ge, Yuying and
             Ge, Yixiao and Shan, Ying},
  booktitle = {Proceedings of the IEEE/CVF Conference on Computer Vision and Pattern Recognition},
  pages   = {13299--308},
  year    = {2023}
}

@article{chen2024mmstar,
  title   = {Are We on the Right Way for Evaluating Large
             Vision-Language Models?},
  author  = {Chen, Lin and Li, Jinsong and Dong, Xiaoyi and Zhang, Pan and
             Zang, Yuhang and Chen, Zehui and Duan, Haodong and Wang,
             Jiaqi and Qiao, Yu and Lin, Dahua and Zhao, Feng},
  journal = {Advances in Neural Information Processing Systems},
  volume  = {37},
  pages   = {27056--87},
  year    = {2024}
}

@inproceedings{lei2020tvqa,
  title     = {{TVQA+}: Spatio-Temporal Grounding for Video Question Answering},
  author    = {Lei, Jie and Yu, Licheng and Berg, Tamara and Bansal, Mohit},
  booktitle = {Proceedings of the 58th annual meeting of the association for computational linguistics},
  pages   = {8211--25},
  year      = {2020}
}

@inproceedings{xiao2021next,
  title     = {{NExT-QA}: Next Phase of Question-Answering to Explaining
               Temporal Actions},
  author    = {Xiao, Junbin and Shang, Xindi and Yao, Angela and Chua, Tat-Seng},
  booktitle = {Proceedings of the IEEE/CVF conference on computer vision and pattern recognition},
  pages     = {9777--86},
  year      = {2021}
}

@article{mangalam2023egoschema,
  title     = {{EgoSchema}: A Diagnostic Benchmark for Very Long-form Video
               Language Understanding},
  author    = {Mangalam, Karttikeya and Akshulakov, Raiymbek and Malik, Jitendra},
  journal = {Advances in Neural Information Processing Systems},
  volume  = {36},
  pages     = {46212--44},
  year      = {2023}
}

@inproceedings{liu2024tempcompass,
  title   = {{TempCompass}: Do Video {LLMs} Really Understand Videos?},
  author  = {Liu, Yuanxin and Li, Shicheng and Liu, Yi and Wang, Yuxiang and
             Ren, Shuhuai and Li, Lei and Chen, Sishuo and Sun, Xu and Hou, Lu},
  booktitle = {Findings of the Association for Computational Linguistics: ACL},
  pages     = {8731--72},
  year    = {2024}
}

@article{cai2024temporalbench,
  title   = {{TemporalBench}: Benchmarking Fine-grained Temporal Understanding
             for Multimodal Video Models},
  author  = {Cai, Mu and Tan, Reuben and Zhang, Jianrui and Zou, Bocheng
             and Zhang, Kai and Yao, Feng and Zhu, Fangrui and Gu, Jing and
             Zhong, Yiwu and Shang, Yuzhang and Dou, Yao and Park, Jaden and Gao, Jianfeng and Lee, Yong Jae and Yang, Jainwei},
  journal = {arXiv preprint arXiv:2410.10818},
  year    = {2024}
}

@inproceedings{isola2015discovering,
  title     = {Discovering States and Transformations in Image Collections},
  author    = {Isola, Phillip and Lim, Joseph J. and Adelson, Edward H.},
  booktitle = {Proceedings of the IEEE conference on computer vision and pattern recognition},
  pages     = {1383--91},
  year      = {2015}
}

@article{souza2023state,
  title     = {Object State Change Classification in Egocentric Videos using the Divided Space-Time Attention Mechanism},
  author    = {Islam, Md Mohaiminul and Bertasius, Gedas},
  journal = {arXiv preprint arXiv:2207.11814},
  year      = {2022}
}

@article{oprea2020review,
  title   = {A Review on Deep Learning Techniques for Video Prediction},
  author  = {Oprea, Sergiu and Martinez-Gonzalez, Pablo and Garcia-Garcia, Alberto
             and Castro-Vargas, John Alejandro and Orts-Escolano, Sergio and
             Garcia-Rodriguez, Jose and Argyros, Antonis},
  journal = {IEEE Transactions on Pattern Analysis and Machine Intelligence},
  volume  = {44},
  number  = {6},
  pages   = {2806--26},
  year    = {2020}
}

@inproceedings{ning2020torque,
  title     = {{TORQUE}: A Reading Comprehension Dataset of Temporal Ordering
               Questions},
  author    = {Ning, Qiang and Wu, Hao and Han, Rujun and Peng, Nanyun and
               Gardner, Matt and Roth, Dan},
  booktitle = {Proceedings of the 2020 Conference on Empirical Methods in Natural Language Processing (EMNLP)},
  pages     = {1158--72},
  year      = {2020}
}

@inproceedings{saxena2021question,
  title   = {Question Answering Over Temporal Knowledge Graphs},
  author  = {Saxena, Apoorv and Chakrabarti, Soumen and Talukdar, Partha},
  booktitle = {Proceedings of the 59th Annual Meeting of the Association for Computational Linguistics and the 11th International Joint Conference on Natural Language Processing},
  volume  = {1},
  pages     = {6663--76},
  year    = {2021}
}

@inproceedings{park2020visualcomet,
  title     = {{VisualCOMET}: Reasoning about the Dynamic Context of a Still Image},
  author    = {Park, Jae Sung and Bhagavatula, Chandra and Mottaghi, Roozbeh and
               Farhadi, Ali and Choi, Yejin},
  booktitle = {European Conference on Computer Vision (ECCV)},
  pages     = {508--24},
  year      = {2020}
}

@article{wei2022cot,
  title   = {Chain-of-Thought Prompting Elicits Reasoning in Large Language
             Models},
  author  = {Wei, Jason and Wang, Xuezhi and Schuurmans, Dale and Bosma, Maarten
             and Ichter, Brian and Xia, Fei and Chi, Ed H. and Le, Quoc V. and
             Zhou, Denny},
  journal = {Advances in Neural Information Processing Systems},
  volume  = {35},
  pages   = {24824--37},
  year    = {2022}
}

@inproceedings{wang2023self,
  title   = {Self-Consistency Improves Chain of Thought Reasoning in Language
             Models},
  author  = {Wang, Xuezhi and Wei, Jason and Schuurmans, Dale and Le, Quoc
             and Chi, Ed and Narang, Sharan and Chowdhery, Aakanksha and
             Zhou, Denny},
  booktitle = {International Conference on Learning Representations (ICLR)},
  year    = {2023}
}

@inproceedings{gao2023pal,
  title   = {{PAL}: Program-aided Language Models},
  author  = {Gao, Luyu and Madaan, Aman and Zhou, Shuyan and Alon, Uri and
             Liu, Pengfei and Yang, Yiming and Callan, Jaime and Neubig, Graham},
  booktitle = {International conference on machine learning},
  pages   = {10764--99},
  year    = {2023}
}

@article{hu2022lora,
  title   = {{LoRA}: Low-Rank Adaptation of Large Language Models},
  author  = {Hu, Edward and Shen, Yelong and Wallis, Phillip and
             Allen-Zhu, Zeyuan and Li, Yuanzhi and Wang, Shean and
             Wang, Lu and Chen, Weizhu},
  journal = {International Conference on Learning Representations (ICLR)},
  year    = {2022}
}

@article{biderman2024lora,
  title   = {{LoRA} Learns Less and Forgets Less},
  author  = {Biderman, Dan and Portes, Jacob and Ortiz, Jose Javier Gonzalez and
             Paul, Mansheej and Greengard, Philip and Jennings, Connor and
             King, Daniel and Havens, Sam and Chiley, Vitaliy and Frankle,
             Jonathan and Blakeney, Cody and Cunningham, John P.},
  journal = {Transactions on Machine Learning Research},
  year    = {2024}
}

@inproceedings{wolf2020transformers,
  title   = {Transformers: State-of-the-Art Natural Language Processing},
  author  = {Wolf, Thomas and Debut, Lysandre and Sanh, Victor and Chaumond,
             Julien and Delangue, Clement and Moi, Anthony and Cistac, Pierric
             and Rault, Tim and Louf, R{\'e}mi and Funtowicz, Morgan and Davison, Joe and
             others},
  booktitle = {Proceedings of the 2020 conference on empirical methods in natural language processing: system demonstrations},
  pages   = {38--45},
  year    = {2020}
}

@inproceedings{li2021representing,
  title   = {Representing Videos as Discriminative Sub-graphs for Action
             Recognition},
  author  = {Li, Dong and Qiu, Zhaofan and Pan, Yingwei and Yao, Ting
             and Li, Houqiang and Mei, Tao},
  booktitle = {Proceedings of the IEEE/CVF conference on computer vision and pattern recognition},
  pages   = {3310--19},
  year    = {2021}
}

@article{parthaw1,
  title   = {Street Object Detection from Synthesized and Processed Semantic Image: A Deep Learning Based Study},
  author  = {Goswami, Parthaw and Hossain, ABM Aowlad},
  journal = {Human-Centric Intelligent Systems},
  pages   = {487--507},
  year    = {2023}
}

@inproceedings{parthaw2,
  title   = {An end-to-end web-based system for rice leaf disease classification using deep learning},
  author  = {Goswami, Parthaw and Hossain, ABM Aowlad and Sakib, Abu Noman Md},
  booktitle = {International Joint Conference on Advances in Computational Intelligence},
  pages   = {517--31},
  year    = {2022}
}

@inproceedings{parthaw3,
  title   = {Corn Leaf Disease Identification via Transfer Learning: A Comprehensive Web-Based Solution},
  author  = {Goswami, Parthaw and Safi, Abdullah Al and Sakib, Abu Noman Md and Datta, Tirtha},
  booktitle = {International Conference on Sustainable and Innovative Solutions for Current Challenges in Engineering {\&} Technology},
  pages   = {429--41},
  year    = {2023}
}

@inproceedings{parthaw4,
  title   = {PrivEraserVerify: Efficient, Private, and Verifiable Federated Unlearning},
  author  = {Goswami, Parthaw and Islam, Md Khairul and Yeafi, Ashfak},
  booktitle = {2025 28th International Conference on Computer and Information Technology (ICCIT)},
  pages   = {5650--55},
  year    = {2025}
}

@inproceedings{parthaw5,
  title   = {SwinTextUNet: Integrating CLIP-Based Text Guidance into Swin Transformer U-Nets for Medical Image Segmentation},
  author  = {Yeafi, Ashfak and Goswami, Parthaw and Islam, Md Khairul and Shamme, Ashifa Islam},
  booktitle = {2025 28th International Conference on Computer and Information Technology (ICCIT)},
  pages   = {4260--65},
  year    = {2025}
}

@article{parthaw6,
  title   = {KIRA: Knowledge-Intensive Image Retrieval and Reasoning Architecture for Specialized Visual Domains},
  author  = {Goswami, Parthaw and Deep, Jaynto Goswami},
  journal = {arXiv preprint arXiv:2604.16915},
  year    = {2026}
}

@article{parthaw7,
  title   = {Image Denoising: A Comprehensive Review of Classical to Deep Learning Approaches},
  author  = {Goswami, Parthaw and Islam, Md Khairul},
  journal = {researchgate},
  year    = {2026}
}

\appendix

\section{Prompt Templates}
\label{app:prompts}

\subsection{Direct Prompting}

\subsubsection*{\taskdelta{}}
\begin{verbatim}
You are given two images of the same scene or subject at
different points in time. Estimate the approximate time
elapsed between the two images.

Choose exactly one of: minutes, hours, days, weeks,
months, years.

Respond in JSON only:
{"time_gap": "<your choice>"}
\end{verbatim}

\subsubsection*{\taskrank{}}
\begin{verbatim}
You are given 4 images of the same scene or subject at
different points in time, presented in a shuffled order
(labelled 0-3). Output the indices in chronological
order from earliest to latest.

Respond in JSON only:
{"order": [<index>, <index>, <index>, <index>]}
\end{verbatim}

\subsubsection*{\tasklocate{}}
\begin{verbatim}
You are given a single image of an ongoing process.
Estimate how far along the process is on a 5-point scale.

Choose exactly one of: very_early, early, mid, late,
final.

Respond in JSON only:
{"stage": "<your choice>"}
\end{verbatim}

\subsubsection*{\taskreverse{}}
\begin{verbatim}
You are given 4 images in sequence (labelled 0-3).
Determine whether the sequence proceeds in forward
chronological order or is time-reversed.

Respond in JSON only:
{"direction": "forward"} or {"direction": "reversed"}
\end{verbatim}

\subsubsection*{\taskodd{}}
\begin{verbatim}
You are given 4 images. Three images are from the same
ongoing process, and one image is a temporal outlier
from a different process. Identify the index (0-3) of
the outlier.

Respond in JSON only:
{"odd_index": <0|1|2|3>}
\end{verbatim}

\subsection{Temporal-Cue Prompting (Example: \taskdelta{})}

\begin{verbatim}
You are given two images of the same scene or subject at
different points in time. Before answering, identify at
least two observable visual cues that indicate the
passage of time (e.g., colour changes, texture
degradation, size differences, structural development).

Then estimate the approximate time elapsed between the
two images. Choose exactly one of: minutes, hours, days,
weeks, months, years.

Respond in JSON only:
{
  "reasoning": "<your step-by-step analysis>",
  "time_gap": "<your choice>"
}
\end{verbatim}

\section{Per-Task Metric Derivations}
\label{app:metrics}

\subsection{Kendall's $\tau$ for \taskrank{}}

For four images, there are $\binom{4}{2} = 6$ pairs.
A perfect prediction ($\tau = 1$) has all 6 pairs concordant; a perfect
reverse prediction ($\tau = -1$) has 0 concordant pairs.
Random prediction yields $\mathbb{E}[\tau] = 0$ and
$\text{Var}(\tau) = 2(2n+5)/(9n(n-1)) = 0.241$ for $n=4$.
The standard deviation of $0.53$ is consistent with the empirical results in
Table~\ref{tab:main_direct}.

\subsection{Log-Scale Vocabulary for \taskdelta{}}

The six time buckets and their representative durations are:
\begin{center}
\begin{tabular}{lrl}
\toprule
Bucket & Representative & $\log$ (seconds) \\
\midrule
minutes & 60\,s    & 4.09 \\
hours   & 3{,}600\,s & 8.19 \\
days    & 86{,}400\,s & 11.37 \\
weeks   & 604{,}800\,s & 13.31 \\
months  & 2{,}592{,}000\,s & 14.77 \\
years   & 31{,}536{,}000\,s & 17.27 \\
\bottomrule
\end{tabular}
\end{center}
Adjacent buckets are separated by $1.46$--$4.1$ units on the log-second scale,
ensuring that the vocabulary imposes a meaningful penalty for off-by-one
predictions.

\begin{figure*}[t]
  \centering
  \includegraphics[width=\linewidth]{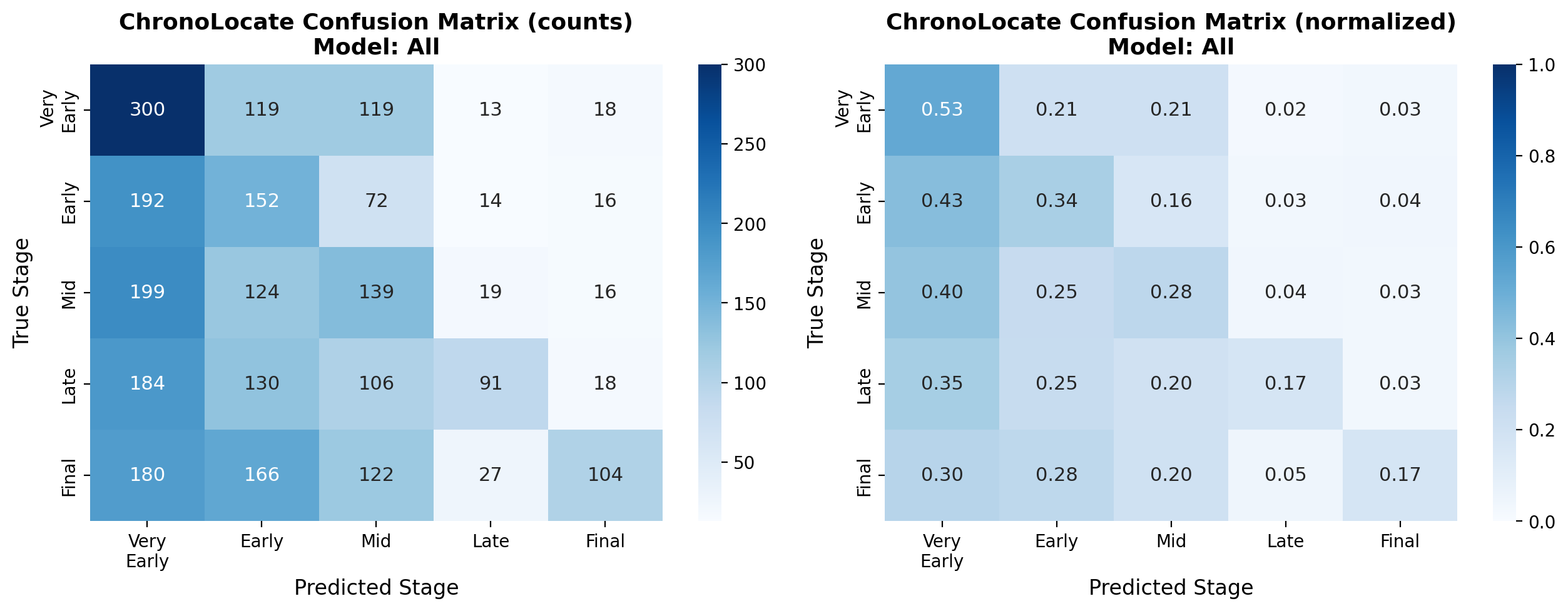}
  \caption{\tasklocate{} confusion matrix aggregated across all VLMs.
           Rows are ground-truth stages; columns are predicted stages.
           Off-diagonal mass indicates the direction and magnitude of
           stage misclassification.}
  \label{fig:locate_confusion}
\end{figure*}

\begin{figure*}[t]
  \centering
  \includegraphics[width=0.95\linewidth]{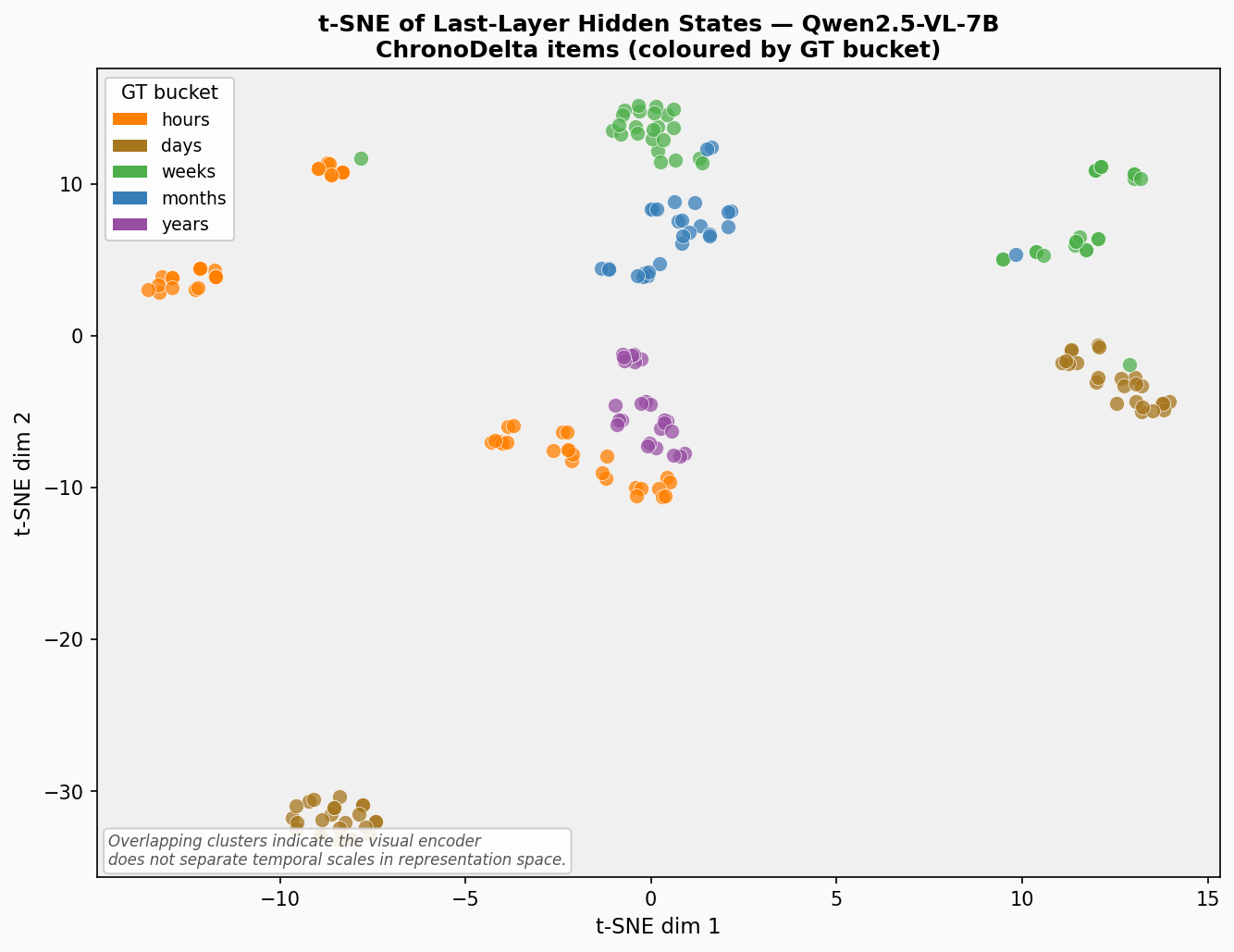}
  \caption{t-SNE projection of the last-layer visual embeddings for
           \taskdelta{} items, coloured by ground-truth time bucket.
           }
  \label{fig:tsne_timebuckets}
\end{figure*}

\begin{figure*}[t]
  \centering
  \includegraphics[width=0.95\linewidth]{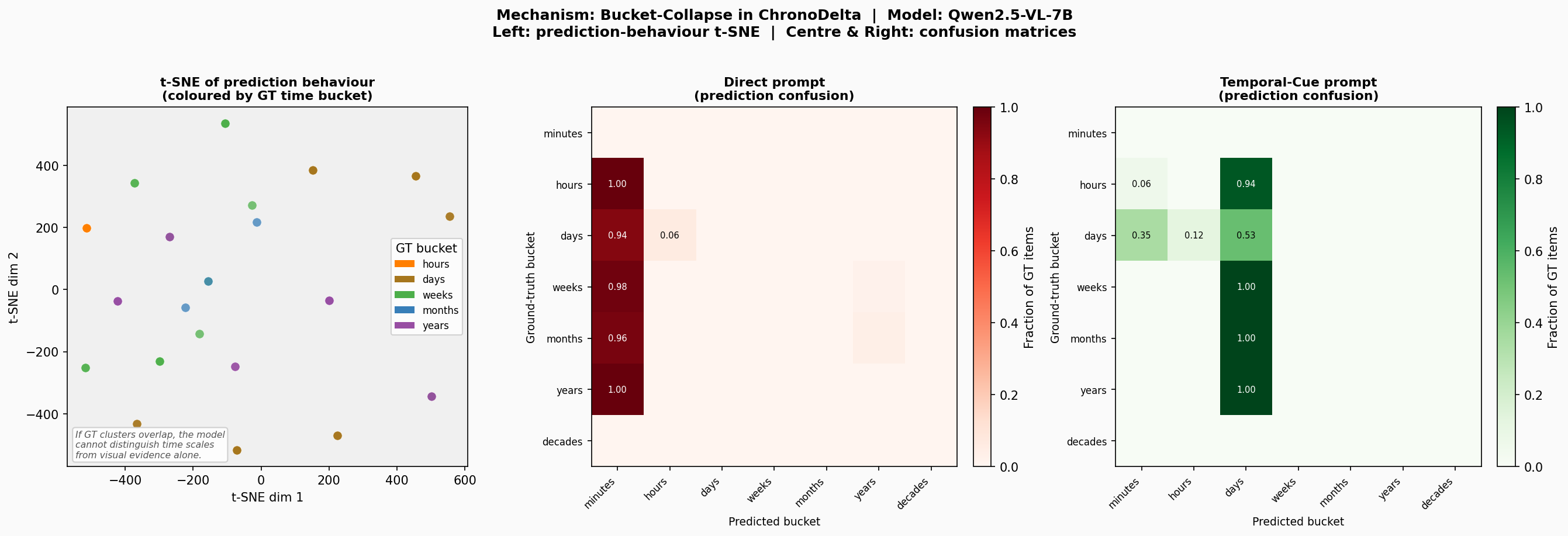}
  \caption{t-SNE projection coloured by the model's predicted time bucket.
           }
  \label{fig:tsne_logits}
\end{figure*}

\begin{figure*}[t]
  \centering
  \includegraphics[width=\linewidth]{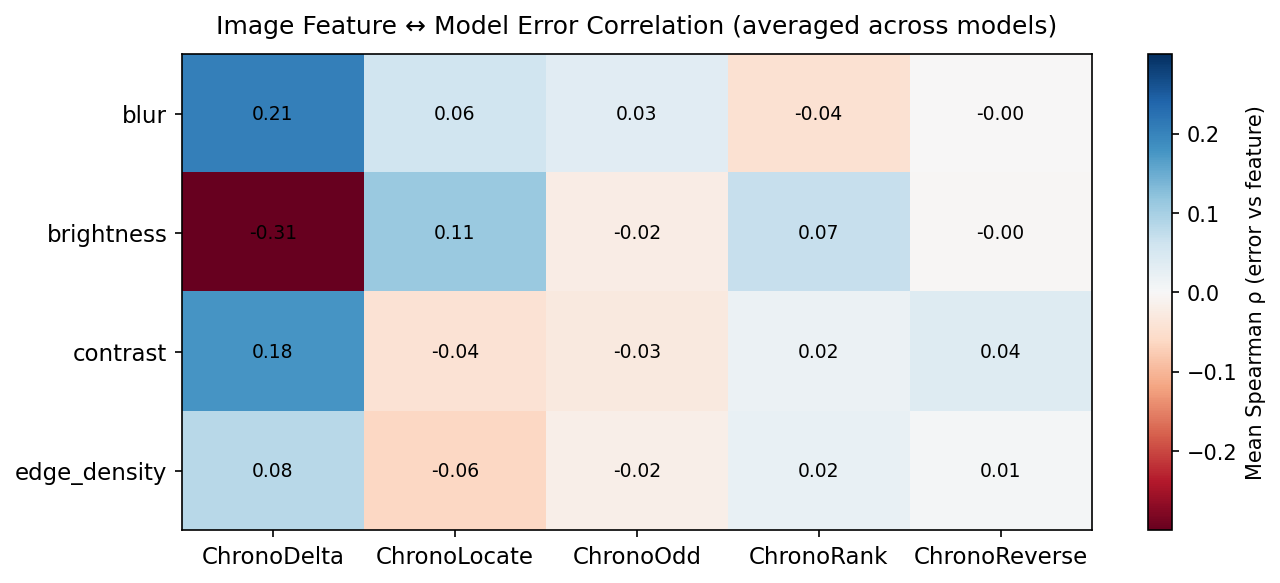}
  \caption{Correlation between model error magnitude and image-level
           visual features (entropy, edge density, colour variance, CLIP
           embedding norm) across items. Error magnitude is
           the log-scale distance between predicted and true time bucket.
           }
  \label{fig:error_corr}
\end{figure*}

\end{document}